\documentclass[10pt,twocolumn,letterpaper]{article}

\usepackage[final]{cvpr}      %

\usepackage{graphicx}
\usepackage{amsmath}
\usepackage{amssymb}
\usepackage{booktabs}
\usepackage{epsfig}
\usepackage{xcolor}
\usepackage{mathtools}
\usepackage{multirow}
\usepackage[linesnumbered,ruled,vlined]{algorithm2e}

\SetCommentSty{mycommfont}
\usepackage[export]{adjustbox}

\definecolor{darkred}{rgb}{0.7,0.1,0.1}
\definecolor{darkgreen}{rgb}{0.1,0.6,0.1}
\definecolor{cyan}{rgb}{0.7,0.0,0.7}
\definecolor{otherblue}{rgb}{0.1,0.4,0.8}
\definecolor{maroon}{rgb}{0.76,.13,.28}
\definecolor{burntorange}{rgb}{0.81,.33,0}

\usepackage[pagebackref,breaklinks,colorlinks]{hyperref}

\usepackage[capitalize]{cleveref}
\crefname{section}{Sec.}{Secs.}
\Crefname{section}{Section}{Sections}
\Crefname{table}{Table}{Tables}
\crefname{table}{Tab.}{Tabs.}

\def\cvprPaperID{1006} %
\def\confName{CVPR}
\def\confYear{2023}

\begin{document}

\title{Prediction of Scene Plausibility}

\author{Or Nachmias\\
Reichman University\\
{\tt\small or.nachmias@post.runi.ac.il}
\and
Ohad Fried \\
Reichman University\\
{\tt\small ofried@runi.ac.il}
\and
Ariel Shamir \\
Reichman University\\
{\tt\small arik@runi.ac.il}
}
\maketitle

\begin{abstract}
Understanding the 3D world from 2D images involves more than detection and segmentation of the objects within the scene. It also includes the interpretation of the structure and arrangement of the scene elements. Such understanding is often rooted in recognizing the physical world and its limitations, and in prior knowledge as to how similar typical scenes are arranged. In this research we pose a new challenge for neural network (or other) scene understanding algorithms - can they distinguish between plausible and implausible scenes? Plausibility can be defined both in terms of physical properties and in terms of functional and typical arrangements. Hence, we define plausibility as the probability of encountering a given scene in the real physical world. We build a dataset of synthetic images containing both plausible and implausible scenes, and test the success of various vision models in the task of recognizing and understanding plausibility. Our source code is available at \url{https://github.com/ornachmias/scene_plausibility}
\end{abstract}

\section{Introduction}
\begin{figure}
\includegraphics[width=.23\textwidth,frame]{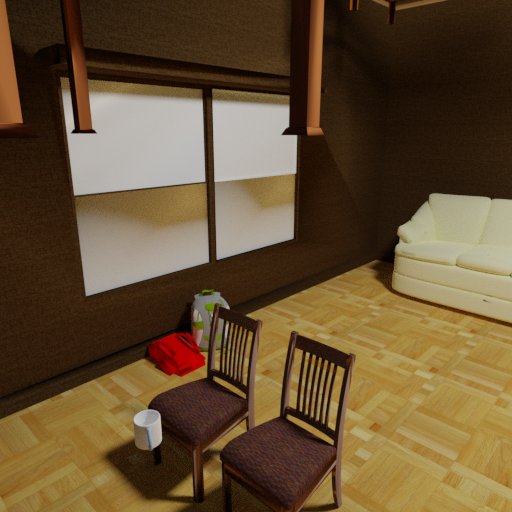}
\includegraphics[width=.23\textwidth,frame]{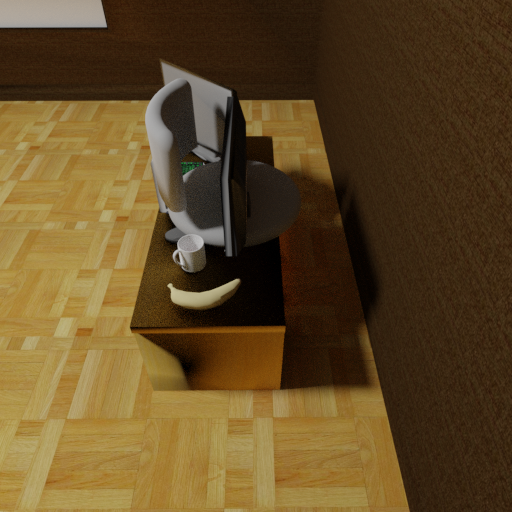}
\\[\smallskipamount]
\includegraphics[width=.23\textwidth,frame]{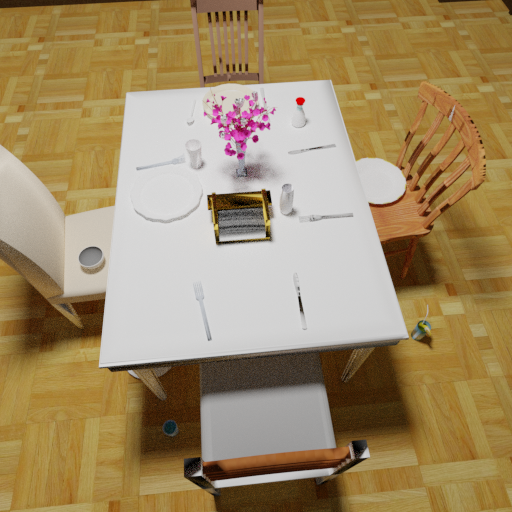}
\includegraphics[width=.23\textwidth,frame]{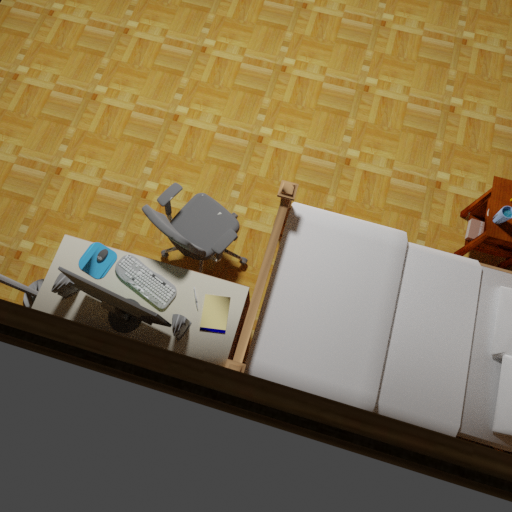}
\\[\smallskipamount]
\includegraphics[width=.23\textwidth,frame]{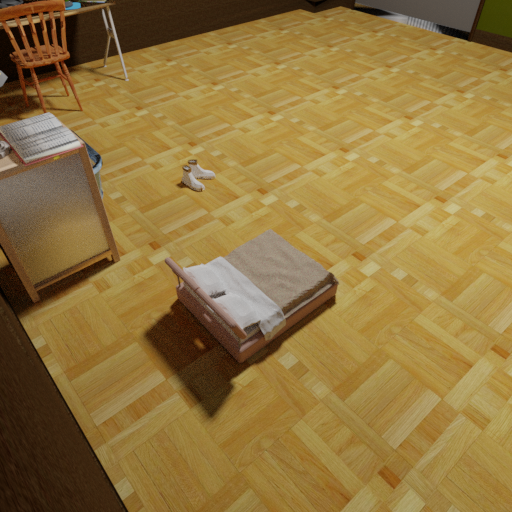}
\includegraphics[width=.23\textwidth,frame]{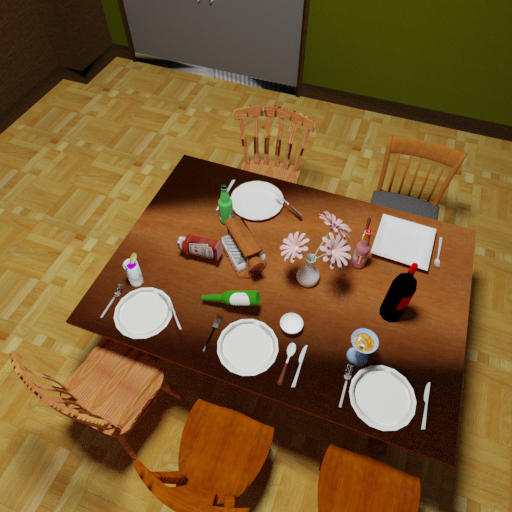}
\caption{Implausible images of different types taken from our database. Top row: gravity violation (left), intersecting objects (right). Middle row: co-occurrence location violations (left), co-occurrence rotation violations (right). Bottom: size inconsistency (left), pose inconsistency (right).}
\label{fig:dataset_examples}
\end{figure}

3D scene understanding from 2D images is a central task in computer vision.
Complete understanding of a scene involves many aspects: one must be able to segment its various parts, detect and understand their meaning and relationships, as well as perceive the general structure of the scene. 
For example, a scene of a home office will typically include a chair, a desk, and perhaps a computer and accessories such as a desk lamp, in a certain arrangement. Any violation of such arrangement (e.g., placing the desk in midair or having the computer monitor face away from the chair) will reduce the plausibility of the scene.
A human can look at such an image and determine whether the scene is plausible or that something is wrong (see examples in Figure~\ref{fig:dataset_examples}).

Only a few computer vision algorithms tackle the task of scene plausibility even though many are proficient in detection and classification of objects and scenes.
In this work, we investigate the notion of plausibility of 2D images of 3D scenes, in terms of arrangement of objects and the general structure of the scene. By leveraging a 3D simulation engine, we create a dataset of plausible and implausible scenes (as defined below). We use this database to measure how well different vision models can classify plausibility, projecting on their ability to understand 3D scene structure as well as the function of the objects within. 

We define plausibility as the probability of encountering a given scene in the real physical world. We focus on indoor scenes of rooms in a house. Some samples in our data are impossible altogether (such as a scene with a flying desk), meaning that their probability is zero, while others are possible, but somehow less probable (such as the turned monitor, or a plate on a dining chair). For this reason we not only define plausibility as a binary attribute but also experiment with a regression model that can rate how plausible a scene is using a value between 0 (implausible) to 1 (plausible). We assume that a scene is plausible only if all of its parts and their arrangement are plausible. Hence, to create implausible scenes we begin with a plausible 3D scene modeled by humans and apply one of a closed set of operation that violate plausibility to a random object in the scene. We define the following set of operations that modify the scene so that it can become implausible (we extend the definition proposed by Bomatter et al.~\cite{bomatter2021pigs} for contextual reasoning in a scene): 

\begin{enumerate}
    \item \textbf{Texture \& color.} Changing the texture or color of an object to be unrealistic, such as a marble plant.
    \item \textbf{Geometry.} Changing the shape of an object so that it is deformed, such as a twisted chair.
    \item \textbf{Gravity.} Changing the position of an object so that it is not supported, such as a desk in the air.
    \item \textbf{Intersection.} Changing the position of an object so that it penetrates or intersects another object.
    \item \textbf{Co-occurrence.} Changing the position of an object to somewhere it is usually not found, such as placing a dinner plate on the floor. 
    \item \textbf{Size.} Changing the size of an object relative to its surrounding objects, such as a huge cup on a dining table.
    \item \textbf{Pose.} Changing the relative positioning of an object to some nonfunctional or improbable configuration, such as placing a chair upside down. 
\end{enumerate}

Since we are interested more in scene structure and relations between objects, we do not include in our data the first two modifications that relate more to the objects themselves and less to the relations in the scene. Note that some modifications such as Gravity and Intersection modification create impossible scenes while all the others can still create physically possible scenes, albeit improbable in the real world. Since the actual probability of the resulting implausible scene is difficult to measure, we approximate the regressive rate of plausibility by measuring the relative size of the modified objects in a given image.

Our work can be seen as following a line of research on the difficulties of vision models on out-of-distribution or adversarial examples~\cite{AdverserialGoodfellow2015,rosenfeld2018elephant,alcorn2019strike,bomatter2021pigs}. However, our primary goal is not just to reveal difficulties but to advance AI algorithms' ability to understand scene structure and spatial relation by introducing a new challenge and dataset. Such understanding can come from prior understanding of the physical world (it is implausible that objects will fly), the functionality of objects (it is implausible that computer screens face the wall) as well as by more statistical inference and co-occurrences (it is unlikely that a plate would be placed on the floor). In our experiments, we found indications that such knowledge is still lacking from current vision models, and this may suggest avenues for future enhancements of these models.

Our main contributions are first, defining a new challenge for computer vision of testing plausibility of scenes; second, building a dataset of images for this challenge which will be openly available for future research; third, testing state-of-the-art algorithms on this challenge and comparing them to human-level understanding.

\section{Related Works}
\paragraph{Single Object Plausibility}
While many works explored the affects of implausible objects on the overall results of different tasks, none have searched for a way to tell if an object is plausible or not. Rosenfeld et al.~\cite{rosenfeld2018elephant} explored common failures of state-of-the art object detectors by transplanting out of context patches in an unexpected way. By doing so, they created an implausible scene, which caused objects’ classification in the scene to fail. Alcorn et al.~\cite{alcorn2019strike} have shown that changing the pose of an object in a scene to less common viewing direction can cause misclassification. Bomatter et al.~\cite{bomatter2021pigs} and Zhang et al.~\cite{zhang2020putting} improve recognition by devising a model that can overcome the implausibility of objects. In contrast to testing detectors and classifier, other works have experimented with placing unrelated objects in a scene~\cite{davenport2007consistency,dvornik2018modeling}, where the objects can be unrelated to one another or the background, and tested how this affects the entire scene classification. Singh et al.~\cite{singh2020don} devised a method to detach the object from the context to improve classification. Although all of these works used implausible scenes, none of them tested the ability to recognize and classify implausibility.

\paragraph{Plausibility Related Datasets}
There have been some datasets that focuses on out-of-context objects placement, such as UnRel \cite{peyre2017weakly} that focuses on real images containing unusual relations between objects, with relatively small size of $\sim$ 1000 images. A human evaluated dataset was also presented in \cite{zhang2020putting}, but it contains only $\sim$ 2500 images and mainly focuses on the dimension of specific object in the image. Cut-and-Paste dataset \cite{dwibedi2017cut} uses a unique augmentation technique that can create out-of-context objects in images, but the pasted object may contain artifacts, which can result in bias toward those artifacts. OCD~\cite{bomatter2021pigs} is a dataset using 35 available 3D rooms, resulting in $\sim$ 16,000 images, across multiple different implausibilities (gravity, co-occurrences and size). We extend their definition and build a larger dataset for other implausibility types as well.

\paragraph{Objects in Scenes}
There is some work on correctly placing objects in scenes. Jian et al.~\cite{jiang2012learning} use a robotic arm to correctly place an object, but the calculations are based on point clouds, under some assumptions on the correct places the object should be. Lin et al.~\cite{lin2018st} present a GAN architecture that focused on placing objects in images, but the main evaluation to the suggested composition was human evaluation. Some additional works regarding image composition focus on a specific set of images, such as faces, streets or baskets \cite{azadi2020compositional}. Van Steenkiste et al.~\cite{van2020investigating}  specifically explored GAN capability of composition, but did so on artificial and not actual scenes.

\paragraph{CNN Texture Bias}
While CNN perform well on image classification tasks, previous work have shown that there is major bias to the textures, meaning layout, structure, and shape play a smaller role in the final classification. Geirhos et al.~\cite{geirhos2018imagenet} and Azad et al.~\cite{azad2021texture} explored this texture bias and ways to reduce it on different tasks. Mohla et al.~\cite{mohla2020cognitivecnn} tried to evaluate the shape-texture bias using a new metric. Another recent paper discovered that although the shapes are encoded in the model, usually the textures have more effect on the final classification~\cite{islam2021shape}. This CNN limitation to learn layout, even when providing the necessary data, prevent it from classifying the layout itself, affecting its ability to distinguish plausible from implausible scene images.

\paragraph{Synthetic Data}
Datasets availability have increased in recent years due to the mass of data needed to train more complex models, such as visual transformers that are reported to use 14 million images to achieve the reported results~\cite{dosovitskiy2020image}. To overcome the difficulty of data collection, it has become common to generate synthetic data using 3D rendering to allow data generation for specific tasks \cite{movshovitz2016useful,richter2016playing,abu2018augmented,sundermeyer2018implicit} such as object viewpoint estimation, classification, detection, segmentation and more.

\section{Dataset Generation}
Classification of the implausibility of indoor scene images is a new task. To experiment with baseline algorithms a novel image dataset must be defined. We create our dataset using 3D synthetically rendered images for several reasons. First, creating implausible scenes (as such) is difficult in the physical world. Using rendering, we can easily transform objects to create implausible scenes with or without physical limitation, and without causing any major texture changes between the original and modified images. Second, using rendered 3D models allows us to have more information regarding the objects in the scene, such as the exact geometry, class and visibility of an object. Third, we are more interested in the scene structure and relations between objects than their appearance, so photo-realism is not a top priority in our data, and vision algorithms have shown robustness to synthetically rendered data. 

We built our image dataset using a 3D scenes dataset~\cite{fisher2012example}, containing 65 indoor scenes, which were designed by humans and we assume are plausible. These scenes include living rooms, work spaces, bedrooms and dining rooms, each containing relevant objects for the scene. The scenes were made out of 500 different models and the data contains a single image representing each scene. Our data generation process used these scenes to create over 36,000 images divided by the following plausibility categories: plausible (9004), co-occurrence location (5273), co-occurrence rotation (4902), pose (4588), gravity (5171), size (2408) and intersection (4762). Each scene has between 270 to 1400 images. 
Below we describe our full images generation process, which mostly uses Blender Python API \footnote{\url{https://docs.blender.org/api/2.91/index.html}}.

We categorize the objects in the scenes based on their relative size. Their occurrences in the dataset is such that 11.9\% are large objects, 24.8\% medium and 63.4\% are small. Our generation process took that into account, in order to normalize the sizes of the implausibilities in the different scenes (Section \ref{selecting_objects}).

The dataset was randomly split into train and test sets based on the scenes (train: 55 scenes, test: 10 scenes). We split the dataset based on the scenes to avoid any bias when evaluating the models, and to prevent the models from memorizing the structure of a scene. We publish our split as part of the dataset.

\subsection{Data Enrichment}
The scenes in the original 3D dataset contain information instructing how to properly build the scenes in 3D, and information regarding every model. We added some improvements to allow future research to approach the plausibility problem using different methods, with as few limitations as possible.

\paragraph{Normalizing Models Pose} The imported models are usually delivered in different poses, and then positioned in the scene to fit the scene composition. To allow future work to examine our dataset based on the 3D models instead of just the provided images, we normalized every object in the same class used in our data to the same initial pose (e.g.\ the front of all chairs face the same direction).

\paragraph{Objects Dependency Tree}
Some scenes contain objects such as tables, desks or nightstands that can have several small objects on top of them. If we move or rotate the base objects, it will cause some dependent objects to possibly become implausible without an explicit operation on them. This can cause inconsistencies to the number of implausible objects in the scene we record. To prevent such a case, we transform all dependent objects with the base objects, thus creating a plausible context for the dependent objects while the base object is in an implausible state. To do that, we manually labeled the objects in the scenes in the form of a tree describing which object is on top of which. 

\subsection{Implausible Scenes} 
\label{selecting_objects}
To create implausible scenes we manipulate between 1 to 5 objects in a scene. Our algorithm creates implausible scenes of six types (Section~\ref{sec:operations}).
For each type we build a list of potential objects in the scene to apply the modification by filtering out objects that are irrelevant for the specific transformation (e.g.\ a cup wouldn't be affected by a rotation transformation). We categorize all object classes into a size category: small, medium and large, and aim that each size category will have the same number of transformed objects overall. Towards this end, we add a weight to each potential object in the list to adjust the probability of selecting it based on its size category, so the results will match the desired distribution. 

Next, we iteratively select the the first, second, up to fifth object for each scene at random using the set of weights from the list of potential objects. After selecting each additional object, we search for a valid transformation for the current object according to the implausibility type, define a camera position based on all selected objects so far (from one to five), and test that the rendering will be valid. If the object transformation was completed successfully, we updated the distribution values to decrease the selected object's chance to get re-elected. For each scene we create 6 images - one plausible, and five implausible that modify from 1 to 5 objects in this scene using the same implausibility type (see Section~\ref{sec:rendering}). The full procedure is described in Algorithm~\ref{algo:find_implausibility}.

\begin{figure}
\includegraphics[width=.23\textwidth,frame]{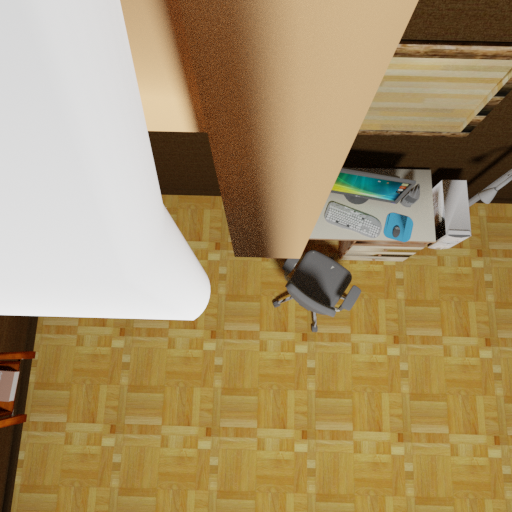}
\includegraphics[width=.23\textwidth,frame]{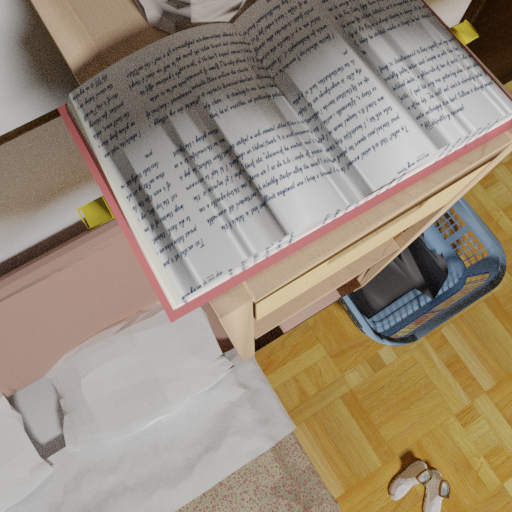}
\caption{Examples of images that were found fit by our algorithm, but contain irregularities which may affect their classification.}
\label{fig:bad_generation_example}
\end{figure}

While the random image generation process allows us to theoretically create an unlimited size dataset based on a relatively small amount of 3D scenes, there are some disadvantages. When we find a camera placement that meets our criteria it can sometimes include unwanted objects that block most of our view (Figure \ref{fig:bad_generation_example}). Such images may affect machine learning models as well as humans, and may affect results by being very easy to detect. Still, we have found that such images are relatively rare.

\begin{algorithm}
\DontPrintSemicolon
\KwIn{$S$: scene in 3D format \\
$T$: current transformation type implementation \\
$N_c$: number of transformations for single camera}
\KwOut{$C_{metadata}$: created camera metadata \\
$C_{trans}$ found transformations visible from camera}

$O \gets []$\ \tcp{Objects vector}
$W \gets []$\ \tcp{Weights vector}
\For{$obj$ in $S.objects$}{
    \If{$isTransformationAllowed(obj, T.type)$}{
        $O.add(obj)$\;
        
        \tcp{Get number of transformed objects across categories}
        $total \gets countTransformedObjects()$
        
        \tcp{Get number of transformed objects in category}
        $categoryTotal \gets countTransformedCategory(obj)$
        
        \tcp{Get target distribution for category}
        $targetDist \gets getTargetDistCategory(obj)$
        
        \tcp{Set weight based on the difference in distributions}
        $currDist \gets categoryTotal / total$\;
        $weight \gets max(0, targetDist - currDist)$
        $W.add(weight)$
    }
}

\tcp{Fallback in case there are not enough weighted objects in the scene}
\If{$len(weights > 0) < N_c$}{
    $W = getTargetDistCategory(O)$
}

$C_{trans} \gets []$

$selectedObjs = randomWeightedChoices(O, W))$

\For{$obj$ in $selectedObjs$}{
    \If{$obj$ in $C_{trans}$}{
        $continue$\;
    }

    \tcp{Find suitable transformation and camera}
    $T_{params} \gets T.find(obj)$
    $C_{metadata} \gets findCamera(T_{params}, C_{trans})$

    \tcp{Validate that the camera matches the transformations}
    \If{$T.validate(T_{params}, C_{trans}, C_{metadata})$}{
        $C_{trans}.add(T_{params})$
    }
    
    \If{$len(C_{trans}) \ge 5$}{
        $break$\;
    }
}

\Return{$C_{metadata}, C_{trans}$}
\caption{Finding implausible transformation of a scene}
\label{algo:find_implausibility}
\end{algorithm}

\subsection{Camera Placement \& Rendering}
\label{sec:rendering}

Given a 3D scene, one can render many different images by choosing different camera viewpoints. Since the original dataset was provided with a single image for each scene, we generated additional images using a stochastic algorithm for camera placement. This algorithm can theoretically allow the creation of any number of unique camera positions, and is only limited by the amount of time the user will allow it to run.

We randomly sample the initial position of the camera in an area above the current transformations and direct it towards the centroid of all original and transformed objects (from 1 to 5). Next, we find a good view by iterating along the vector between the centroid and the original camera location, each time from a different direction of this vector. In each iteration we validate the view by making sure that both the transformed objects and the original objects are visible in the frame and that the camera has a minimum number of objects in view. If the validation fails, we continue the search until we find a valid view. We implemented an early stop mechanism that stops the search in a specific direction if we detect occlusion from this direction, since changing the distance of the camera will keep the same occlusion. The full algorithm is described in Algorithm~\ref{algo:camera_generation}.

Once we find a valid position for all transformations on the same camera, the scene is rendered up to 6 times on that camera; the first render where nothing is transformed and the scene is plausible, and then each render transforms one additional object, creating one plausible image and up to 5 implausible images. Each image is 512x512 pixels, and has a matching metadata file describing the objects’ categories, bounding boxes, visibility by the camera and applied transformation with the original locations in 3D. 

\subsection{Objects Implausibility Modifications}
\label{sec:operations}

We used six types of operations on objects to create implausible scenes. Each is implemented in a stochastic fashion to make sure the dataset can be automatically generated, diverse and unbiased. In the following we explain the implementation of each one.

\textbf{Gravity:} elevate the object in the Z axis by a random factor 1 to 2 times the maximal dimension of the object.

\textbf{Intersection:} we randomly move the object in the X and Y directions by half of the objects' dimensions in the same axis. Additionally, we lower the object randomly by $\frac{1}{3}$ to $\frac{2}{3}$ of its dimension in the Z axis. While the Z axis translation ensure we get intersection, we are using X, Y translation to create a more diverse intersections.

\textbf{Pose:} randomly rotate the object by a random degree in all axes. Then, lower the object so that its lowest point will be in the same height of its lowest point before the transformation, thus avoid overlapping with the gravity category.

\textbf{Size:} randomly scale the object in a random factor either in the range: $[0.3, 0.5]$ (scale down), or in the range $[2, 3]$ (scale up). Then, elevate the object by half of its height, and apply gravity to let it fall into reasonable position. 

\textbf{Co-occurrence (location):} randomly move the object to a random location in a sphere of radius 60 around the original location. Then, apply gravity to the object to arrive at a physically realistic position. If the new object’s height (z-coordinate) is less than $\epsilon$ different from its original, we assume it is still on the same plane and did not really change location (for example, if we move a plate and it remained on the table), so we try again.

\textbf{Co-occurrence (rotation):} rotate the object around the Z axis at a random angle between 160 and 200 degrees. Additionally, randomly move the object in the X and Y axes for up to half the dimension of the matching axis to avoid intersections. If the object intersects with another object, move the object away from the intersecting point until there is no overlap.

\begin{algorithm}
\DontPrintSemicolon 
\KwIn{
$S$: scene in 3D format \\
$T_{params}$: transformed objects information \\
$N_{iter}$: max number of algorithm iterations \\
$step$: size of the camera movement each iteration \\
$L_{cen}$: centroid of the set of original and transformed objects}
\KwOut{Camera parameters: $C_{location}, C_{target}$}

$C_{target} \gets L_{cen}$

\tcp{Generate random location above the centroid}
$C_{location} \gets randomLocation(L_{cen})$

$v \gets buildVector(L_{cen}, C_{location})$

$i \gets 0$

\tcp{Initialize the early stop flags}
$directions = \{ -1: False, 1: False\}$

\While{$i \le N_{iter} \cdot 2$}{
    \For{$d$ in $directions.keys()$}{
        \If{$!directions[d]$}{
            $continue$
        }
        $C_{location} \gets L_{cen} + (v \cdot (\lfloor i/2 \rfloor + 1) \cdot d \cdot step)$
        
        \tcp{Check visibility of object before and after transformation}
        \If{$checkVisibility(S, T_{params})$}{
            \Return{$C_{location}, C_{target}$}
        }
        \If{$checkObscuration(S, T_{params})$}{
            $directions[d] \gets True$
        }
    }
}

\Return{$None, None$}\;
\caption{Camera placement after object's transformation}
\label{algo:camera_generation}
\end{algorithm}

\section{Experimental Setup}
\label{sec:setup}

Our goal is to investigate how well current vision models handle the new implausibility task. For this we chose to compare three well know models: ResNet, ViT and CRTNet. Each model was evaluated in 3 tasks: 
\begin{enumerate}
\setlength{\itemsep}{0pt}
\setlength{\parskip}{0pt}
    \item Classifying plausible from implausible images
    \item Classifying the different types of implausibilities
    \item Providing a regression score to a scene based on the size of the implausibilities in the image.
\end{enumerate}

\subsection{Evaluated Tasks}
\label{sec:tasks}

\textbf{Binary Classification:} Each model was trained and evaluated on a balanced set of plausible and implausible images, while tasked to differentiate between the two. The implausible images were randomly selected from our dataset based on different implausibilities types, and on different numbers of transformed objects to eliminate any kind of bias, both in category or between categories. 

\textbf{Multi-class Classification:} Each model was trained on 7 classes: plausible images, and 6 different categories of implausible images with the objective of differentiating between the classes. Here we sampled the same number of images from the plausible images and each implausibility category.

\textbf{Regression:} We trained the models based on ``plausibility score'', which we define as a continuous value representing how much of the image is implausible. The plausibility score of an image is defined as $1 - \bigcup _{i=1}^{n} B_{i}$,
where $n$ is the number of transformed objects and $B_{i}$ is the bounding box for object $i$ in the image.

\subsection{Models}
\label{sec:models}

Each model was trained for 100 epochs with early stop configured if the validation metrics were not improving for 20 epochs. The training data size was set according to each task: in the classification tasks the categories were equalized, and in the regression task we manually examined the distribution of regression scores and chose training and test samples accordingly. The images that were used were randomly selected in each experiment from the $~36,000$ images in our dataset.

\textbf{ResNet101 \& ViT:} Until recently, CNNs (e.g.\ ResNet) were the state-of-the-art on image classification tasks, but recently models based on visual attention achieved comparable and even better results. While CNNs require less computational resources and less data to train, ViT attention mechanism allows encoding of the relative position of the different features, which on our task may be crucial.
We modified and tested both models in the same way using a pretrained model: for ResNet we used pretrained version based on ImageNet1K \footnote{\url{https://pytorch.org/vision/main/models/generated/torchvision.models.resnet101.html}} and for ViT we used pretrained version based on ImageNet 2012 \footnote{\url{https://github.com/rwightman/pytorch-image-models}}. We then changed the last fully connected layer into the number of classes needed for each task: 2, 7 and 1. Then, we run a full training procedure on all layers to finetune the model. 

\textbf{CRTNet:} The purpose of this model was to correctly classify objects in a scene, even when the object is out of context. CRTNet uses object’s bounding box to create a cropped image, and embedd it together with the full image to provide an accurate prediction. Although the task of this model is different from ours, CRTNet has shown good results by relying on the local and global context of objects, which is very relevant to our task. Even though other models we tested were not provided with the objects' bounding box information, we wante to test this architecture on our tasks as well. Hence, we used CRTNet pretrained on OCD dataset, with an ``oracle'' bounding box information for the transformed objects and trained the classifier and target classifier fully connected layers based on our data.

\section{Results}
\label{sec:results}

\subsection{FID Evaluation}
\label{sec:fid}

Our first objective was to test if the distribution of the plausible and implausible images in our dataset is similar in terms of their appearance and content. One of the tools that are used to identify generated images from real images is the Frechet Inception Distance \cite{heusel2017gans} (FID). While mostly being used to evaluate GANs \cite{heusel2017gans,karras2017progressive,karras2019style}, the method can be used to compare the statistics of two sets of images to see if they come from similar distribution or not. Although some problems were reported with the original FID score \cite{chong2020effectively,shmelkov2018good}, it is still the most commonly used evaluation metric to detect synthetic images.

To compare the distributions of the implausible images and the plausible images, we used publicly available FID calculation \cite{Seitzer2020FID}. Our dataset was split based on the scenes into 4 parts: halving the plausible and implausible scenes in such a way that no part will include the same scenes. We then performed 4-way comparisons between the four sets each containing 4300 images selected randomly based on the splitting criteria. This experiment was conducted 10 times and the average score for each comparison is provided in Table \ref{tab:fid}.

Comparing the FID scores it can be seen that the sets of plausible and implausible images seem to come from the same distribution. 
These lack of differences show that although FID is popular among image generation tasks, it cannot differentiate the context of objects and structure of scenes correctly. 

\begin{table}[t]
\centering
\begin{adjustbox}{width=0.6\columnwidth}
\begin{tabular}{c|c|c}
\toprule
            & Plausible & Implausible \\
\midrule
Plausible   & 38.6        & 39.25         \\ 
\midrule
Implausible & 38.76        & 35.56          \\ 
\bottomrule
\end{tabular}
\end{adjustbox}
\caption{Comparison of FID score between plausible and implausible image sets. }
\label{tab:fid}
\end{table}

\subsection{Human Evaluation}
\label{sec:human}

As an initial baseline we wanted to know how well humans perform in the task of recognizing implausibility on our image dataset. We used an Amazom Mechanical Turk survey on 210 randomly selected images from our database. Each survey contained 6 images and an explanation about the task, where one of the images (a very evident one) acts as an integrity test. We explained the task and asked the workers to perform our multi-class classification tasks (see Section~\ref{sec:tasks}). 

The results for the human evaluation can be seen in Table \ref{tab:human_evaluation}. We present both accuracies by categories and binary classification based on the same answers, which were calculated by using any implausible category selected as implausible. These results indicate that the multi-class classification task is also not simple also for humans, but may also indicate that AMT is not the best platform to run such a test (for example, we had to filter around 70\% of the answers of AMT workers due to failure in the integrity test). Although there were inconsistent results among the different categories of implausibility, we see that humans can recognize scene implausibility with high accuracy ($92\%$), but that it is more difficult to name the exact implausibility type.

\begin{table}[]
\centering
\begin{tabular}{lc}
\toprule
Plausible & 0.63 \\ %
Size & 0.73 \\ %
Gravity & 0.76\\ %
Intersection & 0.62 \\ %
Pose & 0.56 \\ %
\begin{tabular}[c]{@{}c@{}}Co-occurrence Location\end{tabular} & 0.63 \\ %
\begin{tabular}[c]{@{}c@{}}Co-occurrence Rotation\end{tabular} & 0.80 \\ %
\midrule
\begin{tabular}[c]{@{}c@{}}Binary Classification\end{tabular} & 0.92 \\ %
\bottomrule
\end{tabular}
\caption{Accuracy of the human evaluation by category submitted by workers on AMT.}
\label{tab:human_evaluation}
\end{table}

\subsection{Models Performance}
\label{sec:models_perf}

The basic results of our experiments on the three baseline models and three tasks are shown in Table~\ref{tab:models_score}. While all models had partial success in the different tasks, none managed to predict with high accurately when the images are implausible. CRTNet exceeded the other models, but this advantage can be attributed to the additional bounding box information, which focuses the model's resources on a specific object in the scene. Surprisingly, the ViT model that is based on attention to different parts of the image scored lowest on all tasks.

Comparing all models to human binary classification accuracy, it is clear that there is room for improvement. In the multi-class classification task the humans outperformed both ResNet and ViT by a large margin, and had comparable results to CRTNet, but it had the additional `oracle' information of bounding boxes around the location of implausibilities.

\begin{table}
\centering
\begin{adjustbox}{width=0.9\columnwidth}
\begin{tabular}{lccc}
\toprule
& \begin{tabular}[c]{@{}c@{}}Binary \\ Classification\end{tabular} & \begin{tabular}[c]{@{}c@{}}Multi-Class \\ Classification\end{tabular} & \begin{tabular}[c]{@{}c@{}}Plausibility\\ Score\end{tabular} \\
\midrule
ResNet & 0.69 & 0.50 & 0.11 \\ %
ViT & 0.67 & 0.32 & 0.10 \\ %
CRTNet & 0.77 & 0.64 & 0.08  \\ %
\bottomrule
\end{tabular}
\end{adjustbox}
\caption{Comparison of the results of the chosen models for each task. For classification the reported metric is accuracy, and for the plausibility score the reported score is L1 Loss.}
\label{tab:models_score}
\end{table}

\begin{figure*}[ht]
\centering
\includegraphics[width=0.95\textwidth]{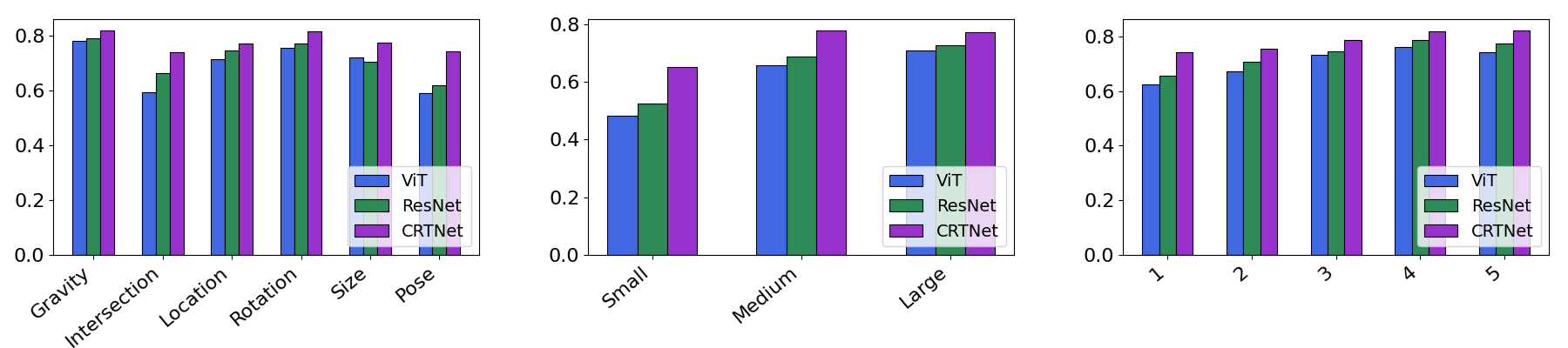}
\caption{Evaluation (accuracy) of the binary classification models based on different subsets of the dataset. From left: implausibility type, objects' relative sizes and number of transformed objects.}
\label{fig:bc_evaluation}
\end{figure*}

To analyze the models’ capabilities better we performed several projections of the results to different subsets. In Figure~\ref{fig:bc_evaluation} we split the binary classification results by implausibility type, number of transformed objects and the relative size of the objects, while keeping the ratio equal between plausible and implausible images in each category. 
When separating according to category of implausibility, it can be seen that the lowest accuracy scores are intersections and pose modifications. Both of these modifications are keeping a relatively similar local context, which may be harder for the algorithms to detect. 
The best accuracy was achieved in detecting gravity change or rotation of an object, which usually affects the local context of the object. 
Relating to size, the differences are not large but an expected trend can be deducted: the smaller the object the more difficult it is to detect its implausibility. Note that actually CRTNet performs slightly worse on the largest objects, possibly because of the local context used in the architecture.
The number of transformed objects had a smaller than expected effect on the accuracy $(<0.1)$, but still, the more objects are transformed, the easier it is for algorithms to figure out that the scene is implausible.

\begin{figure}
\centering
\includegraphics[width=0.4\textwidth]{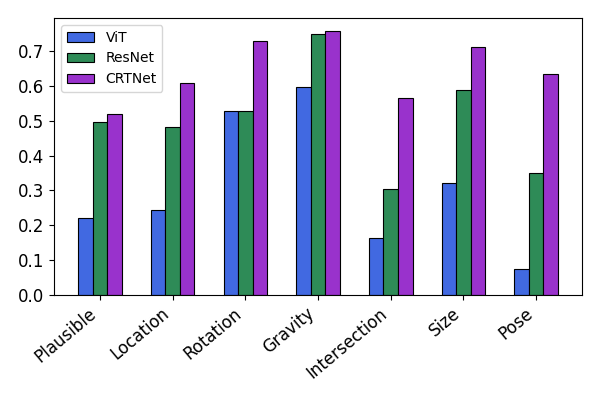}
\caption{Evaluation (accuracy per class) of the multi-class classification results. Each reported result is based on samples in the specific category.}
\label{fig:mcc_evaluation}
\end{figure}

For the multi-class task of different implausibilities type, the models’ performance decreases substantively (Figure \ref{fig:mcc_evaluation}). None of the models managed to classify correctly plausible scenes in more than half of the samples, where some models managed to score even less. The only exception is the gravity implausibility, that was recognized by all models in more than half of the samples. In this task, while CRTNet remained superior in performance, ResNet outperformed ViT by a large margin in most of the categories. While the ability to classify correctly plausible from implausible was mediocre, we can see in this experiment that the models cannot distinguish the type of implausibility with high accuracy. 

\begin{table}[]
\centering
\begin{adjustbox}{width=0.7\columnwidth}
\begin{tabular}{l|c|c|c|c|c}
\toprule
ViT & 0.24 & 0.08 & 0.06 & 0.05 & 0.08 \\ \midrule
ResNet & 0.23 & 0.07 & 0.05 & 0.06 & 0.09 \\ \midrule
CRTNet & 0.18 & 0.07 & 0.05 & 0.05 & 0.05 \\ 
\bottomrule
\end{tabular}
\end{adjustbox}
\caption{Regression score average L1 loss on each 20\% percentile of our dataset, divided by the ground truth plausibility score.}
\label{tab:reg_scores}
\end{table}

Our regression task showed large improvement compared to the classification tasks, where the best result is an average of 0.08 loss on the evaluation set using CRTNet (last column in Table~\ref{tab:models_score}). Further exploration shows that although the scores distribution is similar between our predictions and the ground truth, the model did not learn the actual values for each image. We performed an additional analysis: splitting the dataset into 5 equal parts based on percentiles of ground truth plausibility score, and calculated the average L1 Loss for each part separately. As can be seen in Table~\ref{tab:reg_scores}, low plausibility scores failed to be predicted accurately, and the model only managed to predict correctly based on the mass of the distribution. Moreover, although 7\% of the samples have a plausibility score of 1.0 (plausible), the models scores were 0.2\% (ViT), 2\% (ResNet) and 9\% (CRTNet).

\section{Discussion}
We have presented a new challenge for computer vision algorithms: recognizing plausibility of 3D scenes. Plausibility relates more to the structure of the scene and to relations of objects within it, and less to the individual objects' shape and appearance. We rendered a benchmark dataset and tested the performance of state-of-the-art neural networks on the task. 

It is clear that this task is still a challenge for computer vision algorithms, and we believe that further investigation of this problem will
produce machine learning models that
understand the world around us better, both in terms of its physical properties and in terms of the functionality and typical arrangements of objects. 

Our dataset will be released for future research. In addition, our code is available and can be used to generate additional data with little effort, and can be easily modified to create data of different types --- for example, combining multiple implausibility categories in one image, creating more photo-realistic images, and adding other types of scenes beyond indoor. We hope this manuscript will be a first step towards robust and extensive scene plausibility research.

{\small
\bibliographystyle{ieee_fullname}
\bibliography{main}
}

\clearpage
\begin{center}
\textbf{\large Prediction of Scene Plausibility --- Supplementary Material}
\end{center}
\setcounter{equation}{0}
\setcounter{section}{0}
\setcounter{figure}{0}
\setcounter{table}{0}
\setcounter{page}{1}
\makeatletter
\renewcommand{\theequation}{S\arabic{equation}}
\renewcommand{\thefigure}{S\arabic{figure}}
\renewcommand{\thetable}{S\arabic{table}}
In this supplementary file, we provide additional information about our dataset and further analysis of the models' results. This allows us to compare the distribution of our data against the predictions of the models, thus looking for any bias or memorization performed by the models.

\section{Dataset Distribution}
\label{dataset_dist}
The original dataset contained 26 objects' types, one of which is ``uncategorized'', used for different items that did not fall under any other type. We extended this definition into 47 objects' types by re-labeling all 3D models in the dataset. While we kept the ``uncategorized'' category, we minimized it significantly. In the statistics we present, all objects' types that have 0.1\% or less appearances (compared to the total number of objects in the scenes) were removed.

\Cref{tab:categories_dist} shows the distribution of objects’ types in the original Scenes 3D dataset, the occurrences in our plausible and implausible images, and the percent of each object’s type within each implausibility class. We split the objects by the relative size of the model that we assigned when annotating the data, and sort it by the number of appearances in the different scenes. 

In \Cref{tab:categories_imp_dist} we present the relative amount of each object’s type in every specific transformation type. The empty cells indicate that the specific type was not allowed in the specific transformation class, usually due to inconsistencies or irrelevance. For example, if we use co-occurrence location transformation class on ``silverware'' type, usually the object will not be visible because of its relative small size, or if we use co-occurrence rotate transformation class on ``plate'' type, no visible change will be detected due to the plate round shape. We can see the ``chair'', ``table'' and ``monitor'' categories take part in a relatively high number of transformations compared to their appearances in the dataset. This is caused due to  their relatively large size, allowing better visibility after performing a transformation, thus resulting in a valid transformation. It is important to observe that even though there is a higher number of transformations using  these types, the number of implausible images containing them remains similar to the original distribution, due to images containing multiple instances of those objects, meaning that the dataset still has a high variety of objects.

\section{Models' Predictions}
\label{models_preds}
\begin{figure*}
\centering
\includegraphics[width=.3\textwidth]{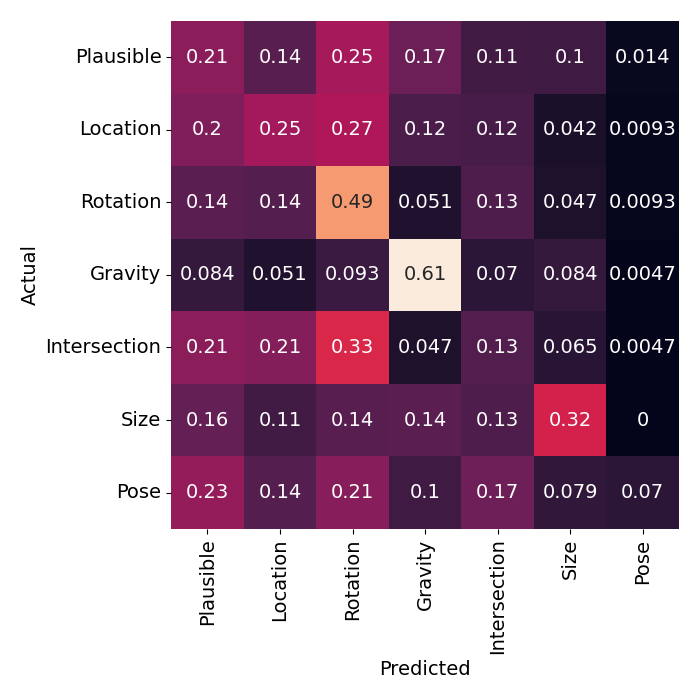}
\includegraphics[width=.3\textwidth]{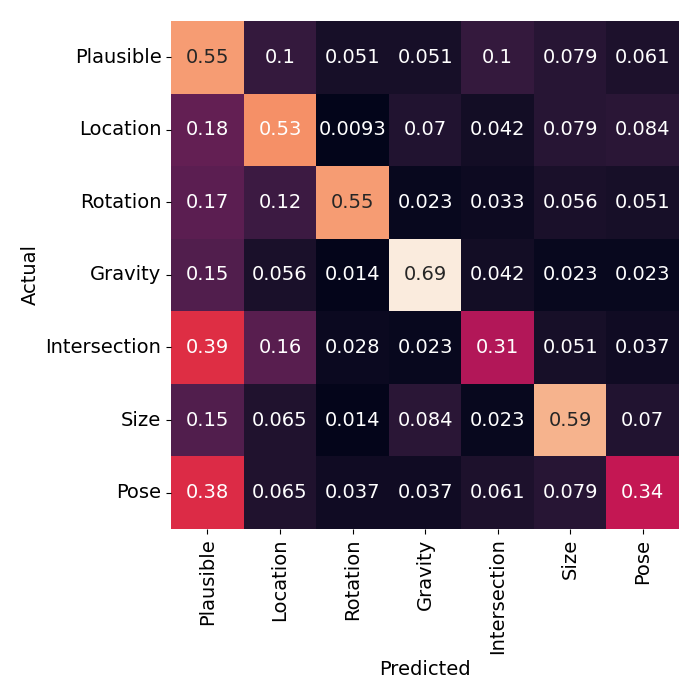}
\includegraphics[width=.3\textwidth]{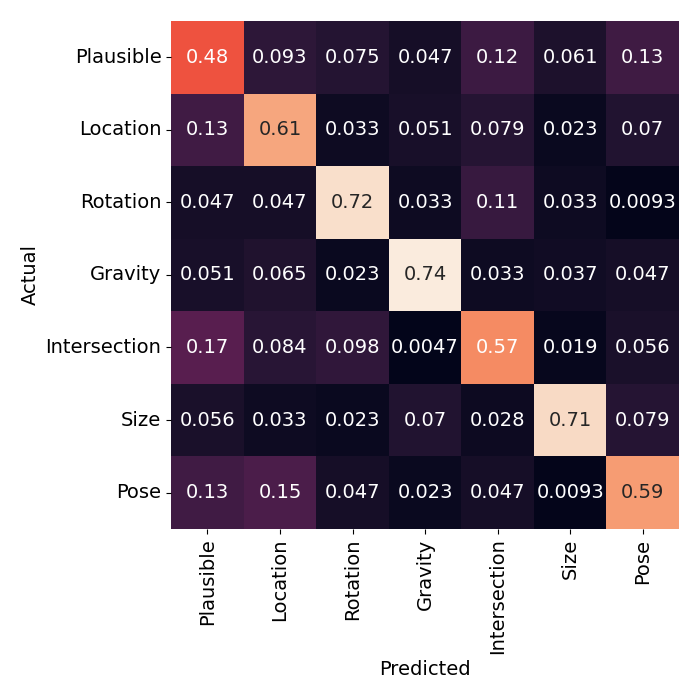}
\caption{The confusion matrices of the evaluated models for the multi-task classification of implausibility class. From left: ViT, ResNet, CRTNet.}
\label{fig:cf_models}
\end{figure*}

To check for any biases in the models’ predictions, we created a subset of our dataset containing implausible images and their matching plausible images. We then ran the images through both the binary classification and classification of implausibilities on all models. We counted the correct prediction in each output class based on the transformed objects’ type, and evaluated the results based on different metrics. In \Cref{fig:cf_models} we can see the confusion matrices for the implausibilities classification for each model, which shows that there is no obvious bias toward any one of the implausibility classes. In \Cref{tab:cf_models_categories} we categorized the binary classification results based on the objects’ types, in an attempt to find objects that have been consistently categorized into one of the plausibility classes. In one specific occurrence the ``mousepad'' object's type was marked as implausible consistently by ViT, but there are no additional cases.

In the binary classification results that are presented in \Cref{tab:bc_dist}, we can see some implausibility bias toward object types such as ``bottle'', ``plant'', ``bottle'', ``cup'', ``silverware'', ``plate'' and ``notebook'', although none of them has a major part in the transformation distribution. An object type such as ``chair'', which was used in half of all ``co-occurrence rotation'' transformations, was correctly classified in 0.6, 0.69 and 0.83 prediction appearances in the different models. ``Silverware'' object type, which was third of all ``pose'' transformations, scored correctly in this class only 0.08, 0.32 and 0.76.

Nonetheless, in \Cref{tab:mcc_vit_dist,tab:resnet_vit_dist,tab:crt_vit_dist}, which present the accuracy of each implausibility classification separately. We tested for object types that managed to get accuracy above 0.9 for specific transformations, but less than 0.2 on the plausible classification, meaning the model can ``memorized'' the object type and mark it as implausible, but we found no occurrences matching this criteria. 

\begin{figure}[h]
\centering
\includegraphics[width=.4\textwidth]{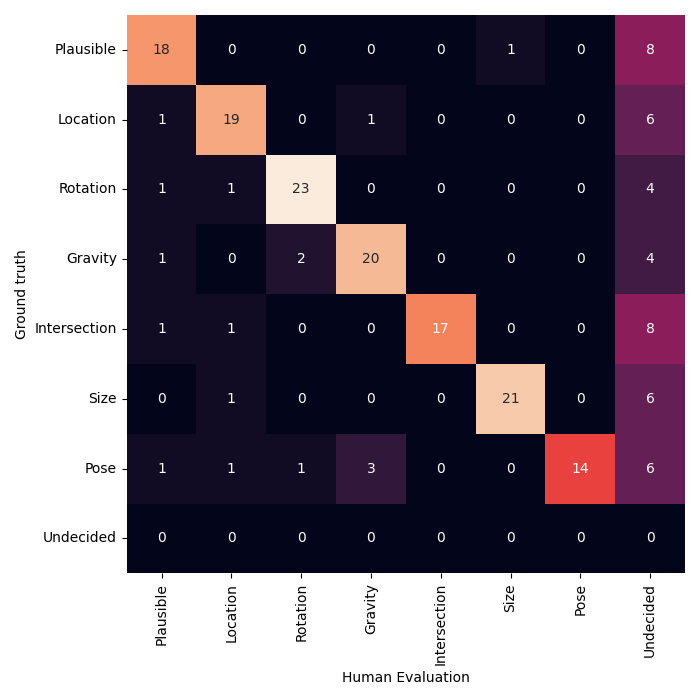}
\caption{The confusion matrix of human evaluation based on majority vote answers. Additional ``Undecided'' row and column were added when more than one class received the votes' majority.}
\label{fig:cf_human}
\end{figure}

\section{Human Evaluation}
\begin{figure*}[h]
\centering
\includegraphics[width=.23\textwidth,frame]{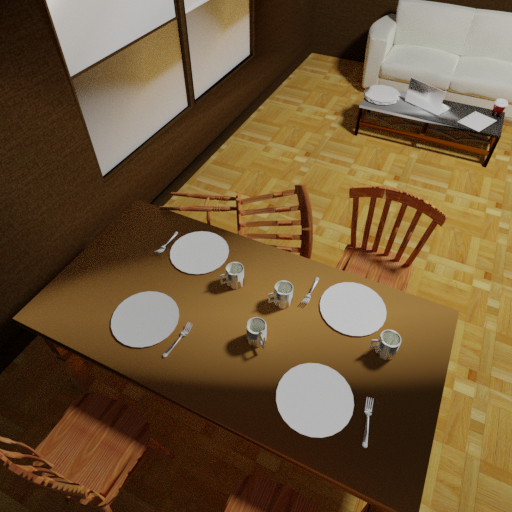}
\includegraphics[width=.23\textwidth,frame]{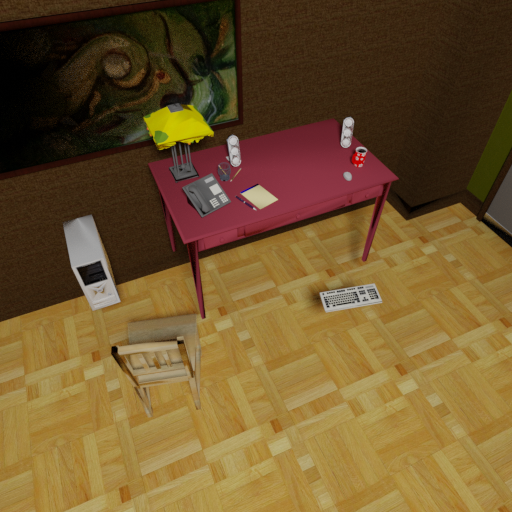}
\includegraphics[width=.23\textwidth,frame]{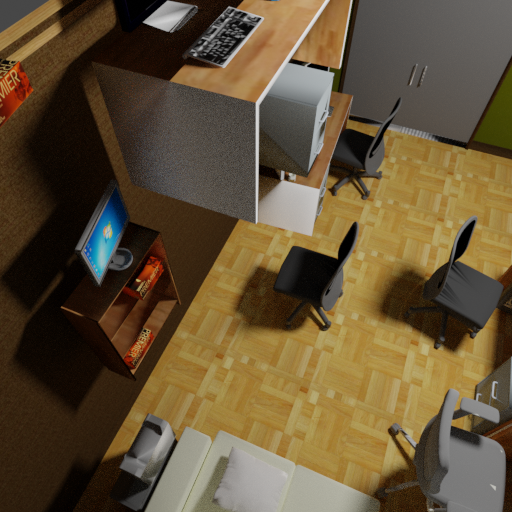}
\includegraphics[width=.23\textwidth,frame]{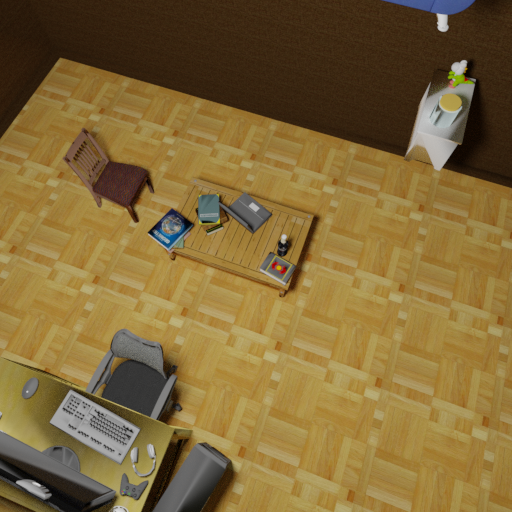}
\\[\smallskipamount]
\includegraphics[width=.23\textwidth,frame]{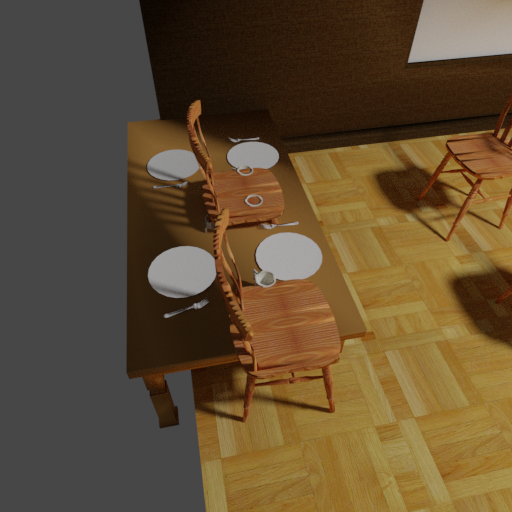}
\includegraphics[width=.23\textwidth,frame]{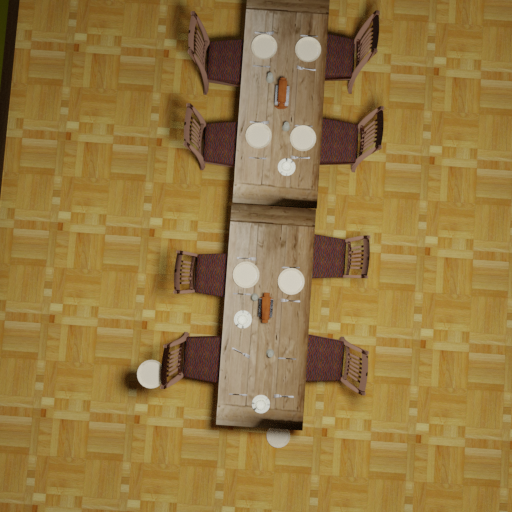}
\includegraphics[width=.23\textwidth,frame]{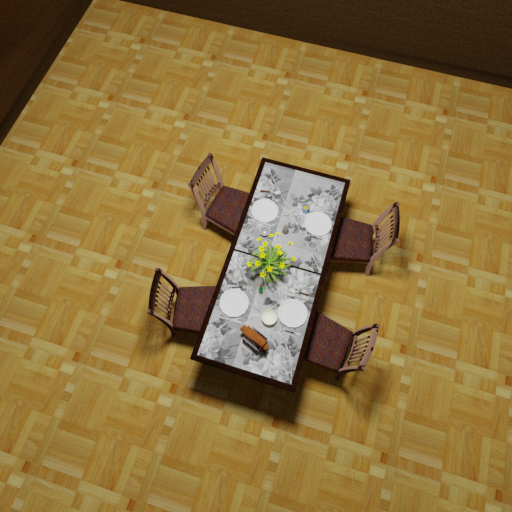}
\includegraphics[width=.23\textwidth,frame]{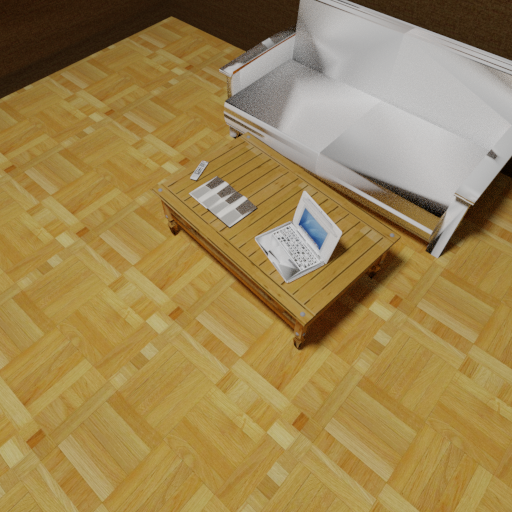}
\caption{The images in our dataset vary in complexity; while some implausibilities are obvious, others are more subtle. Can you spot whether the images are plausible or not? If implausible, try to determine the implausibility type. Images taken from our human evaluation. 
\rotatebox[origin=c]{180}{
\begin{minipage}[t]{\linewidth}
\scriptsize Correct answers are (top left to bottom right): pose (chair), 
co-occurrence location (keyboard, chair, etc.), 
gravity (table, PC, etc.), gravity (right side table), intersection (chairs \& table), 
co-occurrence location (plate), plausible, co-occurrence rotation (laptop)
\end{minipage}
}
}
\label{fig:human_evaluation_images}
\end{figure*}

We published our human evaluation survey on Amazon Mechanical Turk, where each survey contained 6 images. The users were asked to classify each image to one of 7 classes: plausible, or any one of the implausible categories. One of the images in the survey was used as an integrity test --- the answer was fairly obvious and failure to answer it correctly disqualified the entire survey. Out of the 499 surveys published, only 156 passed the integrity test. Each image published received between 1 and 8 answers, and when we evaluated the results we used a majority vote based on the different answers; ties count as an incorrect classification. In addition to the multi-class classification, we aggregated the same answers into ``plausible'' or ``implausible'' and reported the results. 
\Cref{fig:human_evaluation_images} shows several of the images that were used during the study.
In \Cref{fig:cf_human} the confusion matrix of the human evaluation can be seen, including the ``Undecided'' class for cases where the majority vote was split between more than one class. It can be seen that the human evaluation performed better than any of the tested models.

\begin{table*}[]
\centering
\begin{tabular}{lccccc}
\toprule
Object Type & Size & \% Scenes & \% Plausible  & \% Implausible  & \% Transformations \\
\midrule
table & large & 6.9 & 7.4 & 7.3 & 17.8 \\
closet & large & 2 & 1.9 & 1.6 & 4.7 \\
couch & large & 1.7 & 1.7 & 1.5 & 1.9 \\
bed & large & 0.8 & 0.9 & 0.6 & 1.1 \\
\midrule
chair & medium & 11 & 12.1 & 13.3 & 21.4 \\
monitor & medium & 5.1 & 4.6 & 4 & 9.3 \\
desktop & medium & 2.3 & 2.1 & 1.9 & 3.1 \\
walldecoration & medium & 1.2 & 1.2 & 1.1 & 1.4 \\
bag & medium & 1 & 0.8 & 0.7 & 1 \\
pillow & medium & 0.9 & 0.7 & 0.8 & 1.2 \\
books & medium & 0.6 & 0.6 & 0.5 & 0.6 \\
basket & medium & 0.4 & 0.4 & 0.3 & 0.2 \\
box & medium & 0.4 & 0.3 & 0.2 & 0.2 \\
picture & medium & 0.3 & 0.3 & 0.2 & 0.4 \\
tv & medium & 0.3 & 0.2 & 0.2 & 0.3 \\
rug & medium & 0.1 & 0.2 & 0.1 & 0 \\
board & medium & 0.1 & 0.1 & 0 & 0 \\
\midrule
silverware & small & 15.4 & 16.6 & 19.1 & 9 \\
plate & small & 10.2 & 11.1 & 12.4 & 7.8 \\
cup & small & 5.9 & 5.3 & 5.6 & 3.5 \\
keyboard & small & 3.7 & 3.4 & 2.9 & 2.3 \\
speakers & small & 3.3 & 3.1 & 2.7 & 1.5 \\
mouse & small & 3.2 & 2.8 & 2.4 & 1.4 \\
book & small & 3.1 & 2.7 & 2.3 & 2.2 \\
lamp & small & 2.8 & 2.7 & 2.3 & 2 \\
notebook & small & 2.2 & 2.2 & 1.9 & 0.9 \\
bottle & small & 1.7 & 1.7 & 1.7 & 0.8 \\
laptop & small & 1.5 & 1.5 & 1.3 & 0.7 \\
pen & small & 1.2 & 1.4 & 1.2 & 0.5 \\
mousepad & small & 1.2 & 1 & 0.8 & 0.5 \\
control & small & 1 & 0.9 & 0.8 & 0.6 \\
plant & small & 0.7 & 0.7 & 0.7 & 0.3 \\
headphones & small & 0.6 & 0.5 & 0.4 & 0.2 \\
shoes & small & 0.5 & 0.5 & 0.4 & 0.3 \\
sodacan & small & 0.4 & 0.3 & 0.3 & 0.1 \\
tissuebox & small & 0.4 & 0.2 & 0.2 & 0.1 \\
cam & small & 0.3 & 0.1 & 0.1 & 0.1 \\
shirt & small & 0.3 & 0.3 & 0.2 & 0.1 \\
toy & small & 0.2 & 0.1 & 0.1 & 0 \\
clock & small & 0.2 & 0.3 & 0.2 & 0.2 \\
pillbottle & small & 0.2 & 0.2 & 0.2 & 0.1 \\
phone & small & 0.2 & 0.2 & 0.2 & 0.1 \\
videogameconsole & small & 0.1 & 0.1 & 0 & 0 \\
camera & small & 0.1 & 0.1 & 0 & 0 \\
fruit & small & 0.1 & 0.2 & 0.1 & 0 \\
stapler & small & 0.1 & 0.1 & 0.1 & 0 \\
\bottomrule
\end{tabular}
\caption{Objects' types distribution in different contexts, to check for any imbalance in our dataset. The columns describe (from left to right): name of the object's type, assigned relative size category, percent of appearance in the original 3D scenes, number of rendered objects' type divided by total number of rendered objects in the plausible images, number of rendered objects' type divided by total number of rendered objects in the implausible images, and percent of appearance in the total transformations. }
\label{tab:categories_dist}
\end{table*}

\begin{table*}[]
\centering
\begin{tabular}{lcccccc}
\toprule
Object Type & \begin{tabular}[c]{@{}c@{}}\% Co-occurrence\\ Location\end{tabular} & \begin{tabular}[c]{@{}c@{}}\% Co-occurrence\\ Rotation\end{tabular} & \% Pose & \% Size & \% Intersection & \% Gravity \\
\midrule
table & 22.7 & 23.1 & 7.8 & 12.4 & 14.5 & 21.6 \\
closet & 5.6 & 3.2 & 0.9 & 3.3 & 7.2 & 6.5 \\
couch & - & 2.9 & 0.6 & 5.9 & 1.5 & 3.4 \\
bed & 1.7 & - & 0.5 & 2.9 & 0.1 & 2.4 \\
\midrule
chair & - & 50.5 & 12.3 & 29.9 & 21.8 & 20.9 \\
monitor & 8.3 & 14.6 & 8.4 & 6.5 & 10.2 & 6 \\
desktop & 3.9 & - & 0.5 & 6.2 & 6.1 & 4.1 \\
walldecoration & 5.3 & - & - & - & 2.1 & - \\
bag & - & - & - & 2.9 & 2.5 & 1.9 \\
pillow & 2.2 & - & - & - & 2.5 & 1.8 \\
books & 0.9 & - & 0.8 & 1.5 & 0.6 & 0.3 \\
basket & 0.3 & - & 0.3 & 0.1 & 0.4 & 0.2 \\
box & - & - & 0.3 & - & 0.6 & 0.4 \\
picture & 0.5 & - & 0.6 & - & 0.9 & - \\
tv & 0.3 & 0.5 & 0.5 & - & 0.3 & 0 \\
rug & - & - & 0 & - & - & 0 \\
board & 0 & - & - & - & - & 0 \\
\midrule
silverware & - & - & 31.7 & 16.4 & - & 12.4 \\
plate & 25.9 & - & 9.1 & - & - & 4.7 \\
cup & 6.7 & - & 3.7 & - & 6 & 2.2 \\
keyboard & 1.8 & 5.1 & 2.4 & 0.3 & 1.9 & 0.8 \\
speakers & 1.8 & - & 2.8 & - & 2.6 & 1 \\
mouse & 1.7 & - & 2.6 & 1.4 & 2.2 & 0.8 \\
book & 3.2 & - & 3.4 & 1 & 2.9 & 2 \\
lamp & 2.6 & - & 1.7 & 1.7 & 2.9 & 2.6 \\
notebook & - & - & 1.8 & 2.5 & 1.7 & 0.6 \\
bottle & 1.2 & - & 0.4 & 1 & 2.4 & 0.3 \\
laptop & 0.9 & - & 1 & 0.3 & 1.2 & 0.3 \\
pen & 0.5 & - & 1.1 & 1.5 & 0.7 & 0.1 \\
mousepad & 0.5 & - & 1.3 & - & 0.4 & 0.3 \\
control & - & - & 1.2 & 0.3 & 1.4 & 0.4 \\
plant & - & - & 0.2 & 0.4 & 1.1 & 0.3 \\
headphones & 0.3 & - & 0.2 & 0.7 & 0.4 & 0.1 \\
shoes & 0.6 & - & 0.5 & 0.2 & 0.1 & 0.5 \\
sodacan & 0 & - & 0.2 & 0.1 & 0.3 & 0.1 \\
tissuebox & 0.1 & - & 0.1 & 0.4 & 0.2 & 0.1 \\
cam & 0.1 & 0 & 0.2 & 0.1 & 0.2 & 0 \\
shirt & - & - & 0.2 & 0 & - & 0.1 \\
toy & 0.1 & - & 0 & 0 & 0.1 & 0 \\
clock & 0.2 & - & 0.2 & 0.1 & 0.1 & 0.3 \\
pillbottle & 0.2 & - & 0 & 0 & 0 & 0.2 \\
phone & 0 & - & 0.2 & 0.1 & 0.1 & 0 \\
videogameconsole & - & - & 0 & 0 & 0 & 0 \\
camera & 0 & 0 & 0 & 0.1 & 0 & 0 \\
fruit & 0 & 0 & 0 & 0 & - & 0 \\
stapler & 0 & - & 0.1 & 0 & 0.1 & 0 \\
\bottomrule
\end{tabular}
\caption{Number of rendered objects' type divided by total number of rendered objects in the implausible class images. Redacted cells indicate that the specific category was not allowed to transform in the specific implausibility class.}
\label{tab:categories_imp_dist}
\end{table*}

\begin{table*}[]
\centering
\begin{tabular}{lcccccc}
\toprule
 & \multicolumn{2}{c}{ViT} & \multicolumn{2}{c}{ResNet} & \multicolumn{2}{c}{CRTNet} \\
 \cmidrule(l{2pt}r{2pt}){2-3} \cmidrule(l{2pt}r{2pt}){4-5} \cmidrule(l{2pt}r{2pt}){6-7}
Object Type & Plausible & Implausible & Plausible & Implausible & Plausible & Implausible \\
\midrule
laptop & 0.296 & 0.704 & 0.704 & 0.296 & 0.426 & 0.574 \\
mousepad & 0.000 & 1.000 & 0.333 & 0.667 & 0.250 & 0.750 \\
notebook & 0.433 & 0.567 & 0.425 & 0.575 & 0.254 & 0.746 \\
toy & 0.423 & 0.577 & 0.731 & 0.269 & 0.192 & 0.808 \\
box & 0.598 & 0.402 & 0.691 & 0.309 & 0.562 & 0.438 \\
bottle & 0.119 & 0.881 & 0.174 & 0.826 & 0.331 & 0.669 \\
desktop & 0.655 & 0.345 & 0.737 & 0.263 & 0.507 & 0.493 \\
couch & 0.495 & 0.505 & 0.691 & 0.309 & 0.415 & 0.585 \\
bed & 0.475 & 0.525 & 0.494 & 0.506 & 0.469 & 0.531 \\
bag & 0.662 & 0.338 & 0.858 & 0.142 & 0.510 & 0.490 \\
plate & 0.349 & 0.651 & 0.408 & 0.592 & 0.477 & 0.523 \\
pillow & 0.680 & 0.320 & 0.455 & 0.545 & 0.365 & 0.635 \\
chair & 0.427 & 0.573 & 0.476 & 0.524 & 0.513 & 0.487 \\
closet & 0.381 & 0.619 & 0.470 & 0.530 & 0.522 & 0.478 \\
control & 0.432 & 0.568 & 0.659 & 0.341 & 0.477 & 0.523 \\
silverware & 0.206 & 0.794 & 0.292 & 0.708 & 0.442 & 0.558 \\
speakers & 0.412 & 0.588 & 0.676 & 0.324 & 0.235 & 0.765 \\
shoes & 0.511 & 0.489 & 0.457 & 0.543 & 0.447 & 0.553 \\
table & 0.538 & 0.462 & 0.615 & 0.385 & 0.486 & 0.514 \\
cup & 0.310 & 0.690 & 0.419 & 0.581 & 0.444 & 0.556 \\
monitor & 0.627 & 0.373 & 0.706 & 0.294 & 0.464 & 0.536 \\
mouse & 0.403 & 0.597 & 0.806 & 0.194 & 0.323 & 0.677 \\
book & 0.417 & 0.583 & 0.436 & 0.564 & 0.452 & 0.548 \\
keyboard & 0.649 & 0.351 & 0.687 & 0.313 & 0.418 & 0.582 \\
plant & 0.156 & 0.844 & 0.365 & 0.635 & 0.281 & 0.719 \\
walldecoration & 0.431 & 0.569 & 0.464 & 0.536 & 0.458 & 0.542 \\
lamp & 0.433 & 0.567 & 0.477 & 0.523 & 0.554 & 0.446 \\
\bottomrule
\end{tabular}
\caption{Binary classification confusion matrix per predicted class, distributed according to the objects' types for each the three tested model.}
\label{tab:cf_models_categories}
\end{table*}

\begin{table*}[]
\centering
\begin{tabular}{lcccccc}
\toprule
 & \multicolumn{2}{c}{ViT} & \multicolumn{2}{c}{ResNet} & \multicolumn{2}{c}{CRTNet} \\
 \cmidrule(l{2pt}r{2pt}){2-3} \cmidrule(l{2pt}r{2pt}){4-5} \cmidrule(l{2pt}r{2pt}){6-7}
Object Type & Plausible & Implausible & Plausible & Implausible & Plausible & Implausible \\
\midrule
desktop & 0.71 & 0.40 & 0.82 & 0.35 & 0.65 & 0.69 \\
pillow & 0.75 & 0.39 & 0.69 & 0.78 & 0.57 & 0.82 \\
chair & 0.70 & 0.85 & 0.82 & 0.87 & 0.86 & 0.85 \\
plant & 0.23 & 0.92 & 0.48 & 0.75 & 0.42 & 0.88 \\
bottle & 0.19 & 0.96 & 0.26 & 0.92 & 0.45 & 0.77 \\
cup & 0.53 & 0.91 & 0.70 & 0.87 & 0.66 & 0.84 \\
silverware & 0.36 & 0.95 & 0.48 & 0.90 & 0.67 & 0.84 \\
closet & 0.67 & 0.91 & 0.77 & 0.83 & 0.85 & 0.86 \\
box & 0.59 & 0.39 & 0.77 & 0.39 & 0.61 & 0.43 \\
table & 0.78 & 0.71 & 0.88 & 0.65 & 0.79 & 0.81 \\
control & 0.55 & 0.68 & 0.73 & 0.41 & 0.59 & 0.64 \\
couch & 0.67 & 0.68 & 0.93 & 0.54 & 0.72 & 0.82 \\
book & 0.69 & 0.86 & 0.78 & 0.90 & 0.74 & 0.85 \\
bed & 0.88 & 0.93 & 0.90 & 0.91 & 0.85 & 0.88 \\
plate & 0.64 & 0.95 & 0.75 & 0.94 & 0.79 & 0.86 \\
lamp & 0.77 & 0.91 & 0.84 & 0.89 & 0.91 & 0.89 \\
walldecoration & 0.63 & 0.77 & 0.73 & 0.80 & 0.66 & 0.81 \\
monitor & 0.75 & 0.50 & 0.87 & 0.46 & 0.71 & 0.77 \\
keyboard & 0.79 & 0.49 & 0.87 & 0.49 & 0.66 & 0.78 \\
mouse & 0.48 & 0.68 & 0.94 & 0.32 & 0.48 & 0.77 \\
bag & 0.74 & 0.41 & 0.88 & 0.17 & 0.67 & 0.65 \\
notebook & 0.51 & 0.64 & 0.55 & 0.70 & 0.40 & 0.85 \\
laptop & 0.41 & 0.81 & 0.93 & 0.52 & 0.74 & 0.74 \\
shoes & 0.77 & 0.74 & 0.83 & 0.91 & 0.74 & 0.79 \\
toy & 0.54 & 0.69 & 0.77 & 0.31 & 0.23 & 0.92 \\
speakers & 0.41 & 0.59 & 0.94 & 0.59 & 0.41 & 0.94 \\
mousepad & 0.00 & 1.00 & 0.33 & 0.67 & 0.33 & 1.00 \\
\bottomrule
\end{tabular}
\caption{Percentage of successfully predicting images as plausible or implausible, based on the transformed objects in the image, to validate if there are any classes associated by the model as either plausible or implausible.}
\label{tab:bc_dist}
\end{table*}

\begin{table*}[]
\centering
\begin{tabular}{lccccccc}
\toprule
Object Type & C0 & C1 & C2 & C3 & C4 & C5 & C6 \\
\midrule
desktop & 0.25 & 0.13 & - & 0.28 & 0.18 & 0.35 & 0 \\
pillow & 0.28 & 0.17 & - & 0.41 & 0.04 & - & - \\
chair & 0.12 & - & 0.6 & 0.77 & 0.19 & 0.44 & 0.06 \\
plant & 0.21 & - & - & 0.2 & 0.19 & - & 0 \\
bottle & 0.22 & 0.32 & - & - & 0.15 & - & - \\
cup & 0.2 & 0.18 & - & 0.75 & 0.13 & - & 0.04 \\
silverware & 0.18 & - & - & 0.54 & - & 0.13 & 0.08 \\
closet & 0.42 & 0.2 & 0.04 & 0.83 & 0.06 & 0.5 & 0 \\
box & 0.11 & - & - & 0.17 & 0.4 & - & 0 \\
table & 0.2 & 0.28 & 0.47 & 0.6 & 0.24 & 0.56 & 0.08 \\
control & 0.36 & - & - & 1 & 0.13 & 1 & 0 \\
couch & 0.33 & - & 0.27 & 0.7 & 0 & 0.7 & - \\
book & 0.54 & 0.05 & - & 0.84 & 0 & 0 & 0 \\
bed & 0.3 & 0.29 & - & 0.82 & 0 & 0.13 & 0 \\
plate & 0.15 & 0.3 & - & 0.61 & - & - & 0.07 \\
lamp & 0.6 & 0.26 & - & 0.82 & 0 & 0 & - \\
walldecoration & 0.41 & 0.25 & - & - & 0 & - & - \\
monitor & 0.19 & 0.17 & 0.37 & 0.25 & 0 & 0 & 0 \\
keyboard & 0.15 & 0 & 0.32 & - & - & - & - \\
mouse & 0.32 & 0.12 & - & 0 & - & 0 & 0 \\
bag & 0.36 & - & - & 0.27 & 0.28 & 0.09 & - \\
notebook & 0.25 & - & - & 0.33 & 0.11 & 0 & 0 \\
laptop & 0.37 & 0 & - & 0.2 & 0 & - & 0 \\
shoes & 0.57 & 0.19 & - & 0.67 & 0 & 0 & 0 \\
toy & 0 & 0.6 & - & 0.5 & 0.5 & - & 0 \\
speakers & 0.47 & 0.06 & - & - & - & - & - \\
mousepad & 0.17 & 0 & - & - & 0 & - & 0 \\
\bottomrule
\end{tabular}
\caption{ViT percentage of successfully predicting images in the correct implausibility class, based on the transformed objects in the image. Implausibility classes: plausible (C0), co-occurrence location (C1), co-occurrence rotation (C2), gravity (C3), intersection (C4), size (C5), pose (C6). 
}
\label{tab:mcc_vit_dist}
\end{table*}

\begin{table*}[]
\centering
\begin{tabular}{lccccccc}
\toprule
Object Type & C0 & C1 & C2 & C3 & C4 & C5 & C6 \\
\midrule
desktop & 0.31 & 0.36 & - & 0.65 & 0.32 & 0.47 & 0.1 \\
pillow & 0.69 & 0.38 & - & 0.81 & 0.25 & - & - \\
chair & 0.67 & - & 0.69 & 0.95 & 0.36 & 0.91 & 0.64 \\
plant & 0.83 & - & - & 0.4 & 0.23 & - & 0.33 \\
bottle & 0.77 & 0.11 & - & - & 0.09 & - & - \\
cup & 0.55 & 0.57 & - & 0.86 & 0.28 & - & 0.36 \\
silverware & 0.73 & - & - & 0.67 & - & 0.21 & 0.32 \\
closet & 0.4 & 0.69 & 0.08 & 0.96 & 0.46 & 0.88 & 0.8 \\
box & 0.13 & - & - & 0.17 & 0.23 & - & 0.08 \\
table & 0.5 & 0.58 & 0.47 & 0.86 & 0.44 & 0.59 & 0.56 \\
control & 0.32 & - & - & 1 & 0.13 & 1 & 0.33 \\
couch & 0.24 & - & 0.2 & 0.9 & 0.5 & 1 & - \\
book & 0.47 & 0.63 & - & 0.93 & 0.34 & 0.25 & 0.19 \\
bed & 0.61 & 0.36 & - & 0.92 & 0.4 & 0.63 & 0 \\
plate & 0.67 & 0.6 & - & 0.72 & - & - & 0.42 \\
lamp & 0.44 & 0.7 & - & 0.91 & 0.33 & 0.25 & - \\
walldecoration & 0.49 & 0.5 & - & - & 0.3 & - & - \\
monitor & 0.33 & 0.67 & 0.41 & 0.5 & 0.2 & 1 & 0.33 \\
keyboard & 0.39 & 1 & 0.34 & - & - & - & - \\
mouse & 0.26 & 0.52 & - & 0 & - & 1 & 0 \\
bag & 0.16 & - & - & 0.33 & 0.26 & 0.55 & - \\
notebook & 0.4 & - & - & 0.33 & 0.29 & 0.75 & 0.19 \\
laptop & 0.37 & 0.67 & - & 0.2 & 0 & - & 0.17 \\
shoes & 0.45 & 0.75 & - & 0.67 & 0.89 & 1 & 0.25 \\
toy & 0.38 & 0.8 & - & 1 & 0 & - & 0 \\
speakers & 0.59 & 0.47 & - & - & - & - & - \\
mousepad & 0.5 & 0 & - & - & 0 & - & 0 \\
\bottomrule
\end{tabular}
\caption{ResNet percentage of successfully predicting images in the correct implausibility class, based on the transformed objects in the image. Implausibility classes: plausible (C0), co-occurrence location (C1), co-occurrence rotation (C2), gravity (C3), intersection (C4), size (C5), pose (C6).}
\label{tab:resnet_vit_dist}
\end{table*}

\begin{table*}[]
\centering
\begin{tabular}{lccccccc}
\toprule
Object Type & C0 & C1 & C2 & C3 & C4 & C5 & C6 \\
\midrule
desktop & 0.44 & 0.18 & - & 0.55 & 0.64 & 0.71 & 0.2 \\
pillow & 0.54 & 0.31 & - & 0.75 & 0.36 & - & - \\
chair & 0.55 & - & 0.83 & 0.89 & 0.64 & 0.91 & 0.77 \\
plant & 0.56 & - & - & 0.2 & 0.32 & - & 0.75 \\
bottle & 0.62 & 0.39 & - & - & 0.19 & - & - \\
cup & 0.56 & 0.66 & - & 0.87 & 0.52 & - & 0.44 \\
silverware & 0.5 & - & - & 0.64 & - & 0.63 & 0.76 \\
closet & 0.54 & 0.8 & 0.2 & 0.94 & 0.56 & 0.5 & 1 \\
box & 0.19 & - & - & 0.14 & 0.33 & - & 0.08 \\
table & 0.49 & 0.63 & 0.65 & 0.85 & 0.69 & 0.69 & 0.54 \\
control & 0.23 & - & - & 1 & 1 & 1 & 0.33 \\
couch & 0.45 & - & 0.38 & 0.9 & 0.75 & 0.9 & - \\
book & 0.53 & 0.79 & - & 0.91 & 0.45 & 0.25 & 0.11 \\
bed & 0.73 & 0.5 & - & 0.98 & 0.2 & 0.88 & 0 \\
plate & 0.56 & 0.73 & - & 0.69 & - & - & 0.52 \\
lamp & 0.54 & 0.8 & - & 0.95 & 0.47 & 0.5 & - \\
walldecoration & 0.56 & 0.54 & - & - & 0.36 & - & - \\
monitor & 0.52 & 0.5 & 0.6 & 1 & 0.2 & 0 & 0.33 \\
keyboard & 0.48 & 0 & 0.52 & - & - & - & - \\
mouse & 0.77 & 0.28 & - & 0 & - & 1 & 0 \\
bag & 0.41 & - & - & 0.53 & 0.9 & 0.59 & - \\
notebook & 0.45 & - & - & 0.33 & 0.46 & 0.5 & 0.19 \\
laptop & 0.26 & 0.22 & - & 0.2 & 0.71 & - & 0.17 \\
shoes & 0.57 & 0.75 & - & 0.78 & 0.56 & 1 & 0.17 \\
toy & 0.23 & 0.2 & - & 1 & 0.5 & - & 0 \\
speakers & 0.82 & 0.12 & - & - & - & - & - \\
mousepad & 0.33 & 0 & - & - & 0.33 & - & 0 \\
\bottomrule
\end{tabular}
\caption{CRTNet percentage of successfully predicting images in the correct implausibility class, based on the transformed objects in the image. Implausibility classes: plausible (C0), co-occurrence location (C1), co-occurrence rotation (C2), gravity (C3), intersection (C4), size (C5), pose (C6).
}
\label{tab:crt_vit_dist}
\end{table*}
\end{document}